\documentclass[runningheads,table]{llncs}
\usepackage{graphicx}
\usepackage{hyperref}

\usepackage{caption}
\usepackage{subcaption}
\usepackage[raggedright]{sidecap}
\sidecaptionvpos{figure}{t} 
\usepackage{multirow}
\usepackage{amsmath}
\usepackage{amssymb}
\usepackage{booktabs}
\usepackage{breqn}
\usepackage{array}
\usepackage{highlight}
\newcolumntype{P}[1]{>{\centering\arraybackslash}p{#1}}
\usepackage{color}
\newcommand\modelname{MDViT}

\begin{document}
\title{\modelname{}: Multi-domain Vision Transformer for Small Medical Image Segmentation Datasets}
%
\titlerunning{\modelname{}}
%
\author{Siyi Du\inst{1}\orcidID{0000-0002-9961-4533} \and
Nourhan Bayasi\inst{1}\orcidID{0000-0003-4653-6081} \and
Ghassan Hamarneh\inst{2}\orcidID{0000-0001-5040-7448} \and
Rafeef Garbi\inst{1}\orcidID{0000-0001-6224-0876}}

%
\authorrunning{Siyi Du et al.}
%
\institute{University of British Columbia, Vancouver, British Columbia, CA \\
\email{\{siyi,nourhanb,rafeef\}@ece.ubc.ca} \\
\and Simon Fraser University, Burnaby, British Columbia, CA \\
\email{hamarneh@sfu.ca}} 
%
\maketitle              
%

\begin{abstract}
Despite its clinical utility, medical image segmentation (MIS) remains a daunting task due to images' inherent complexity and variability. Vision transformers (ViTs) have recently emerged as a promising solution to improve MIS; however, they require larger training datasets than convolutional neural networks. To overcome this obstacle, data-efficient ViTs were proposed, but they are typically trained using a single source of data, which overlooks the valuable knowledge that could be leveraged from other available datasets. Naïvly combining datasets from different domains can result in negative knowledge transfer (NKT), i.e., a decrease in model performance on some domains with non-negligible inter-domain heterogeneity. In this paper, we propose \modelname{}, the first multi-domain ViT that includes domain adapters to mitigate data-hunger and combat NKT by adaptively exploiting knowledge in multiple small data resources (domains).
Further, to enhance representation learning across domains, we integrate a mutual knowledge distillation paradigm that transfers knowledge between a universal network (spanning all the domains) and auxiliary domain-specific network branches. Experiments on 4 skin lesion segmentation datasets show that \modelname{} outperforms state-of-the-art algorithms, with superior segmentation performance and a fixed model size, at inference time, even as more domains are added. Our code is available at \url{https://github.com/siyi-wind/MDViT}.
\end{abstract}

\section{Introduction}\label{section:introduction}
Medical image segmentation (MIS) is a crucial component in medical image analysis, which aims to partition an image into distinct regions (or segments) that are semantically related and/or visually similar. This process is essential for clinicians to, among others, perform qualitative and quantitative assessments of various anatomical structures or pathological conditions and perform image-guided treatments or treatment planning~\cite{asgari2021deep}. Vision transformers (ViTs), with their inherent ability to model long-range dependencies, have recently been considered a promising technique to tackle MIS. They process images as sequences of patches, with each patch having a global view of the entire image. This enables a ViT to achieve improved segmentation performance compared to traditional convolutional neural networks (CNNs) on plenty of segmentation tasks~\cite{han2022survey}. However, due to the lack of inductive biases, such as weight sharing and locality, ViTs are more data-hungry than CNNs, i.e., require more data to train~\cite{touvron2021training}. Meanwhile, it is common to have access to multiple, diverse, yet small-sized datasets (100s to 1000s of images per dataset) for the same MIS task, e.g.,  PH2~\cite{mendoncca2013ph} and ISIC 2018~\cite{codella2019skin} in dermatology, LiTS~\cite{bilic2023liver} and CHAOS~\cite{kavur2021chaos} in liver CT, or OASIS~\cite{marcus2007open} and ADNI~\cite{jack2008alzheimer} in brain MRI. As each dataset alone is too small to properly train a ViT, the challenge becomes how to effectively leverage the different datasets.

\begin{SCtable*}
\centering
\caption{Related works on mitigating ViTs' data-hunger or multi-domain adaptive learning. \textbf{U} (universal) implies a model spans multiple domains. \textbf{F} means the model's size at inference time remains fixed even when more domains are added.}
\label{table:related_works}
\resizebox{0.5\textwidth}{!}{
\begin{tabular}{|p{14mm}|P{7mm}|P{3mm}p{42.5mm}|P{3.8mm}|P{3.8mm}|}
\hline
\textbf{Method} &  \textbf{ViT} &  \multicolumn{2}{c|}{\textbf{Mitigate ViTs' data-hunger}} & \textbf{U} & \textbf{F} \\

\hline
\cite{cao2023swin,zhang2021transfuse,liu2021efficient} & \textcolor{blue}{$\surd$} &   \textcolor{blue}{$\surd$} & by adding inductive bias & \textcolor{red}{$\times$} & - \\
\hline
\cite{touvron2021training,xie2022unimiss} & \textcolor{blue}{$\surd$}  & \textcolor{blue}{$\surd$} & by knowledge sharing& \textcolor{red}{$\times$} & - \\
\hline
\cite{wang2022towards} & \textcolor{blue}{$\surd$} &\textcolor{blue}{$\surd$} & by increasing dataset size &\textcolor{red}{$\times$} & - \\
\hline
\cite{cao2022training}  &\textcolor{blue}{$\surd$}  & \textcolor{blue}{$\surd$} & by unsupervised pretraining &\textcolor{red}{$\times$} & - \\
\hline

\cite{rundo2019use,wang2019towards} & \textcolor{red}{$\times$} & \textcolor{red}{$\times$} & ~ & \textcolor{blue}{$\surd$} & \textcolor{blue}{$\surd$} \\
\hline
\cite{liu2020ms,rebuffi2018efficient}  & \textcolor{red}{$\times$} & \textcolor{red}{$\times$} & ~ & \textcolor{blue}{$\surd$} & \textcolor{red}{$\times$} \\
\hline
\cite{wallingford2022task} & \textcolor{blue}{$\surd$} & \textcolor{red}{$\times$} & ~ & \textcolor{blue}{$\surd$} & \textcolor{red}{$\times$} \\
\hline

\modelname{} & \textcolor{blue}{$\surd$}    & \textcolor{blue}{$\surd$} &by multi-domain learning & \textcolor{blue}{$\surd$} & \textcolor{blue}{$\surd$} \\
\hline
\end{tabular}}
\end{SCtable*}

Various strategies have been proposed to address ViTs' data-hunger (Table~\ref{table:related_works}), mainly: \emph{Adding inductive bias} by constructing a hybrid network that fuses a CNN with a ViT~\cite{zhang2021transfuse}, imitating CNNs' shifted filters and convolutional operations~\cite{cao2023swin}, or enhancing spatial information learning~\cite{liu2021efficient}; \emph{sharing knowledge} by transferring knowledge from a CNN~\cite{touvron2021training} or pertaining ViTs on multiple related tasks and then fine-tuning on a down-stream task~\cite{xie2022unimiss}; \emph{increasing data} via augmentation~\cite{wang2022towards}; and \emph{non-supervised pre-training}~\cite{cao2022training}. Nevertheless, one notable limitation in these approaches is that they are not universal, i.e., they rely on \emph{separate training} for each dataset rather than incorporate valuable knowledge from related domains. As a result, they can incur additional training, inference, and memory costs, which is especially challenging when dealing with multiple small datasets in the context of MIS tasks. Multi-domain learning, which trains a single universal model to tackle all the datasets simultaneously, has been found promising for reducing computational demands while still leveraging information from multiple domains~\cite{adadi2021survey,liu2020ms}. To the best of our knowledge, multi-domain universal models have not yet been investigated for alleviating ViTs' data-hunger.

Given the inter-domain heterogeneity resulting from variations in imaging protocols, scanner manufacturers, etc.~\cite{bayasi2021culprit,liu2020ms}, directly mixing all the datasets for training, i.e., \emph{joint training}, may improve a model's performance on one dataset while degrading performance on other datasets with non-negligible unrelated domain-specific information, a phenomenon referred to as \emph{negative knowledge transfer} (NKT)~\cite{adadi2021survey,zhang2022survey}. A common strategy to mitigate NKT in computer vision is to introduce adapters aiding the model to adapt to different domains, i.e., \emph{multi-domain adaptive training} (MAT), such as domain-specific mechanisms~\cite{liu2020ms,rebuffi2018efficient,wallingford2022task}, and squeeze-excitation layers~\cite{wang2019towards,rundo2019use} (Table~\ref{table:related_works}). However, those MAT techniques are built based on CNN rather than ViT or are scalable, i.e., the models' size at the inference time increases linearly with the number of domains.

To address ViTs' data-hunger, in this work, we propose \modelname{}, a novel fixed-size multi-domain ViT trained to adaptively aggregate valuable knowledge from multiple datasets (domains) for improved segmentation. In particular, we introduce a domain adapter that adapts the model to different domains to mitigate negative knowledge transfer caused by inter-domain heterogeneity. Besides, for better representation learning across domains, we propose a novel mutual knowledge distillation approach that transfers knowledge between a universal network (spanning all the domains) and additional domain-specific network branches.

We summarize our contributions as follows: (1) To the best of our knowledge, we are the first to introduce multi-domain learning to alleviate ViTs' data-hunger when facing limited samples per dataset. (2) We propose a multi-domain ViT, \modelname{}, for medical image segmentation with a novel domain adapter to counteract negative knowledge transfer and with mutual knowledge distillation to enhance representation learning. (3) The experiments on 4 skin lesion segmentation datasets show that our multi-domain adaptive training outperforms separate and joint training (ST and JT), especially a 10.16\% improvement in IOU on the skin cancer detection dataset compared to ST and that \modelname{} outperforms state-of-the-art data-efficient ViTs and multi-domain learning strategies.

\section{Methodology}
\begin{figure}[t]
\centering
\includegraphics[width=0.94\linewidth]{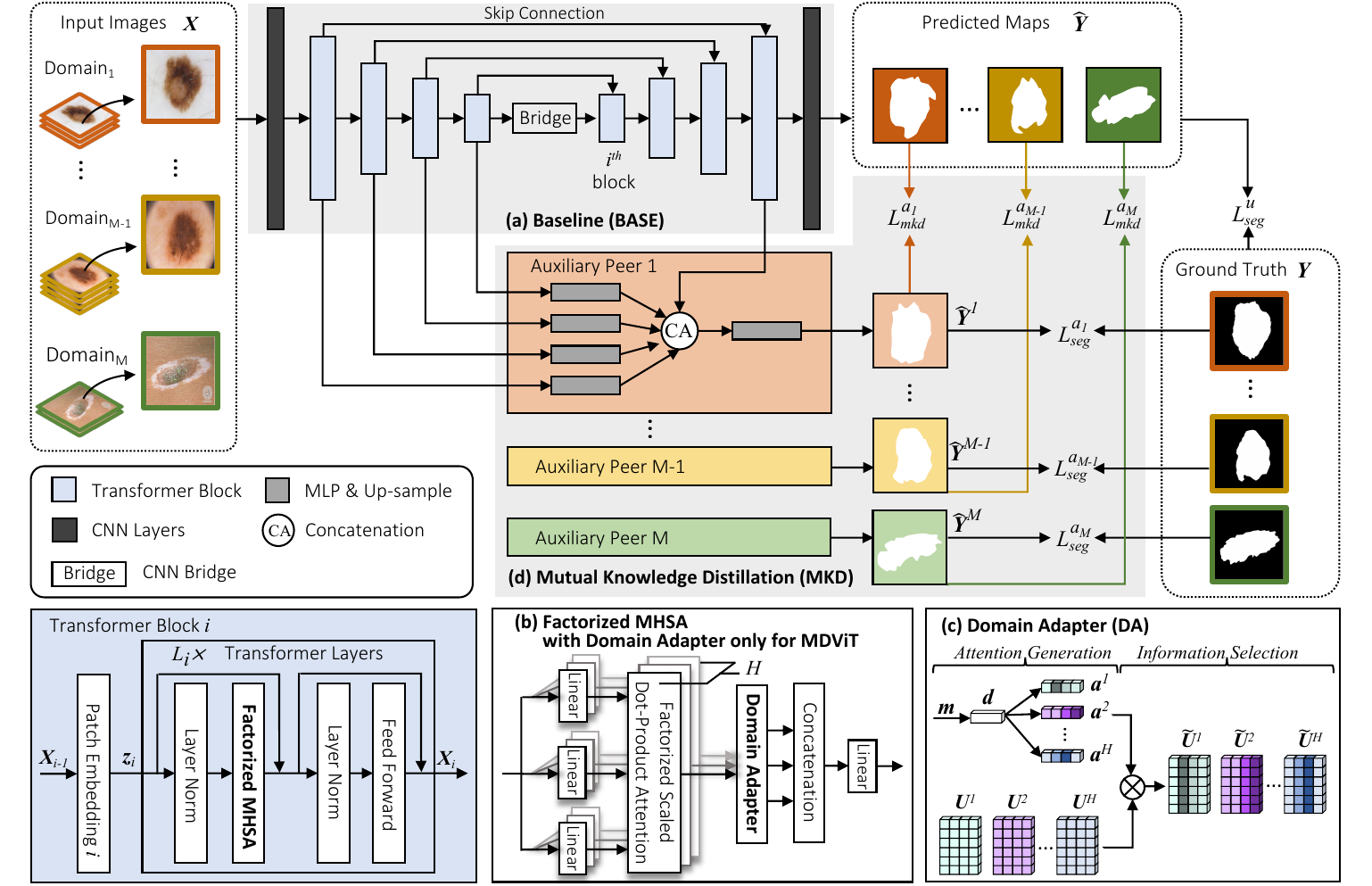}
\caption{Overall architecture of \modelname{}, which is trained on multi-domain data by optimizing two types of losses: $L_{seg}$ and $L_{mkd}$. \modelname{} extends BASE (a) with DA inside factorized MHSA (b), which is detailed in (c), and MKD (d).}
\label{fig:model_overview}
\end{figure}

Let $\boldsymbol{X} \in \mathbb{R}^{H \times W \times 3}$ be an input RGB image and $\boldsymbol{Y} \in \{0,1\}^{H \times W}$ be its ground-truth segmentation mask. Training samples $\{(\boldsymbol{X},\boldsymbol{Y})\}$ come from $M$ datasets, each representing a domain. We aim to build and train a single ViT that performs well on all domain data and addresses the insufficiency of samples in any of the datasets. We first introduce our baseline (BASE), a ViT with hierarchical transformer blocks (Fig.~\ref{fig:model_overview}-a). Our proposed  \modelname{} extends BASE with 1) a domain adapter (DA) module inside the factorized multi-head self-attention (MHSA) to adapt the model to different domains (Fig.~\ref{fig:model_overview}-b,c), and 2) a mutual knowledge distillation (MKD) strategy to extract more robust representations across domains (Fig.~\ref{fig:model_overview}-d). We present the details of \modelname{} in Section~\ref{section:model}.

BASE is a U-shaped ViT based on the architecture of U-Net~\cite{ronneberger2015u} and pyramid ViTs~\cite{lee2022mpvit,cao2023swin}. It contains encoding (the first four) and decoding (the last four) transformer blocks, a two-layer CNN bridge, and skip connections. As described in~\cite{lee2022mpvit}, the $i$th transformer block involves a convolutional patch embedding layer with a patch size of $3\times3$ and $L_i$ transformer layers with factorized MHSA in linear complexity, the former of which converts a feature map $X_{i-1}$ into a sequence of patch embeddings $\boldsymbol{z}_i \in \mathbb{R}^{N_i \times C_i}$, where $N_i=\frac{H}{2^{i+1}} \frac{W}{2^{i+1}}, 1 \leq i \leq 4$ is the number of patches and $C_i$ is the channel dimension. We use the same position embedding as~\cite{lee2022mpvit} and skip connections as~\cite{ronneberger2015u}. To reduce computational complexity, following~\cite{lee2022mpvit}, we add two and one CNN layer before and after transformer blocks, respectively, enabling the 1st transformer block to process features starting from a lower resolution: $\frac{H}{4} \times \frac{W}4$. We do not employ integrated and hierarchical CNN backbones, e.g., ResNet, in BASE as data-efficient hybrid ViTs~\cite{wang2021boundary,zhang2021transfuse}, to clearly evaluate the efficacy of multi-domain learning in mitigating ViTs' data-hunger.

\subsection{\modelname{}}\label{section:model}
\modelname{} consists of a universal network (spanning $M$ domains) and $M$ auxiliary network branches, i.e., peers, each associated with one of the $M$ domains. The universal network is the same as BASE, except we insert a domain adapter (DA) in each factorized MHSA to tackle negative knowledge transfer. Further, we employ a mutual knowledge distillation (MKD) strategy to transfer domain-specific and shared knowledge between peers and the universal network to enhance representation learning. Next, we will introduce DA and MKD in detail.

\noindent\textbf{Domain Adapter (DA):} In multi-domain adaptive training, some methods build domain-specific layers in parallel with the main network~\cite{rebuffi2018efficient,liu2020ms,wallingford2022task}. Without adding domain-specific layers, we utilize the existing parallel structure in ViTs, i.e., MHSA, for domain adaptation. The $H$ parallel heads of MHSA mimic how humans examine the same object from different perspectives~\cite{clark2019does}. Similarly, our intuition of inserting the DA into MHSA is to enable the different heads to have varied perspectives across domains. Rather than manually designate each head to one of the domains, guided by a domain label, \modelname{} learns to focus on the corresponding features from different heads when encountering a domain. DA contains two steps: \emph{Attention Generation} and \emph{Information Selection} (Fig.~\ref{fig:model_overview}-c).

\emph{Attention Generation} generates attention for each head. We first pass a domain label vector $\boldsymbol{m}$ (we adopt one-hot encoding $\boldsymbol{m} \in \mathbb{R}^M$ but other encodings are possible) through one linear layer with a ReLU activation function to acquire a domain-aware vector $\boldsymbol{d} \in \mathbb{R}^{\frac{K}{r}}$. $K$ is the channel dimension of features from the heads. We set the reduction ratio $r$ to 2. After that, similar to~\cite{li2019selective}, we calculate attention for each head: $\boldsymbol{a}^h = \psi(\boldsymbol{W}^h\boldsymbol{d}) \in \mathbb{R}^K, h=1,2,...H$, where $\psi$ is a softmax operation across heads and $\boldsymbol{W}^h \in \mathbb{R}^{K \times \frac{K}{r}}$.

\emph{Information Selection} adaptively selects information from different heads. After getting the feature $\boldsymbol{U}^h = [\boldsymbol{u}_1^h,\boldsymbol{u}_2^h,...,\boldsymbol{u}_K^h] \in \mathbb{R}^{N \times K}$ from the $h$th head, we utilize $\boldsymbol{a}^h$ to calibrate the information along the channel dimension: $\tilde{\boldsymbol{u}}_k^h = a^h_k  \cdot \boldsymbol{u}_k^h$.

\noindent\textbf{Mutual Knowledge Distillation (MKD):} Distilling knowledge from domain-specific networks has been found beneficial for universal networks to learn more robust representations~\cite{liu2020ms,zhou2022novel}. Moreover, mutual learning that transfers knowledge between teachers and students enables both to be optimized simultaneously~\cite{gou2021knowledge}. To realize these benefits, we propose MKD that mutually transfers knowledge between auxiliary peers and the universal network. In Fig.~\ref{fig:model_overview}-d, the $m$th auxiliary peer is only trained on the $m$th domain, producing output $\hat{\boldsymbol{Y}}^m$, whereas the universal network's output is $\hat{\boldsymbol{Y}}$. Similar to~\cite{liu2020ms}, we utilize a symmetric Dice loss $L_{mkd}^{a_m}=Dice(\hat{\boldsymbol{Y}}, \hat{\boldsymbol{Y}}^m)$ as the knowledge distillation loss. Each peer is an expert in a certain domain, guiding the universal network to learn domain-specific information. The universal network experiences all the domains and grasps the domain-shared knowledge, which is beneficial for peer learning.

Each \emph{Auxiliary Peer} is trained on a small, individual dataset specific to that peer (Fig.~\ref{fig:model_overview}-d). To achieve a rapid training process and prevent overfitting, particularly when working with numerous training datasets, we adapt a lightweight multilayer perception (MLP) decoder designed for ViT encoders~\cite{xie2021segformer} to our peers' architecture. Specifically, multi-level features from the encoding transformer blocks (Fig.~\ref{fig:model_overview}-a) go through an MLP layer and an up-sample operation to unify the channel dimension and resolution to $\frac{H}{4} \times \frac{W}{4}$, which are then concatenated with the feature involving domain-shared information from the universal network's last transformer block. Finally, we pass the fused feature to an MLP layer and do an up-sample to obtain a segmentation map.

\subsection{Objective Function}
Similar to Combo loss~\cite{taghanaki2019combo}, BASE's segmentation loss combines Dice and binary cross entropy loss: $L_{seg}=L_{Dice}+L_{bce}$. In \modelname{}, we use the same segmentation loss for the universal network and auxiliary peers, denoted as $L_{seg}^u$ and $L_{seg}^a$, respectively. The overall loss is calculated as follows. 
\begin{equation}\label{equation:loss}
    L_{total} = L_{seg}^u(\boldsymbol{Y}, \hat{\boldsymbol{Y}})+\alpha \sum_{m=1}^M{ L_{seg}^{a_m}(\boldsymbol{Y}, \hat{\boldsymbol{Y}}^m)}+\beta \sum_{m=1}^M{L_{mkd}^{a_m}(\hat{\boldsymbol{Y}}, \hat{\boldsymbol{Y}}^m)}.
\end{equation}
\noindent We set both $\alpha$ and $\beta$ to 0.5. $L_{seg}^{a_m}$ does not optimize DA to avoid interfering with the domain adaptation learning. After training, we discard the auxiliary peers and only utilize the universal network for inference.

\begin{table*}[t]
\centering
\caption{Segmentation results comparing BASE, \modelname{}, and SOTA methods. We report the models' parameter count at inference time in millions (M). \textbf{T} means training paradigms. $^{\dagger}$ represents using domain-specific normalization.}
\label{table:results}
\resizebox{\textwidth}{!}{
\begin{tabular}
{|p{21mm}|p{16mm}|P{9mm}|P{11mm}|P{11mm}|P{11mm}|P{11mm}|P{21mm}|P{11mm}|P{11mm}|P{11mm}|P{11mm}|P{21mm}|}

\hline
\textbf{Model} & \multicolumn{1}{c|}{\textbf{\#Param.}} & \textbf{T} & \multicolumn{10}{c|}{\textbf{Segmentation Results in Test Sets (\%)}} \\
\cline{4-13}
~ & \multicolumn{1}{c|}{\textbf{(millions)}} & ~ & \multicolumn{5}{c|}{\textbf{Dice $\uparrow$}} & \multicolumn{5}{c|}{\textbf{IOU $\uparrow$}} \\
\cline{4-13}
~ & \multicolumn{1}{c|}{\textbf{(M)}} & ~ &  ISIC & DMF & SCD & PH2 & avg $\pm$ std  & ISIC & DMF & SCD & PH2 & avg $\pm$ std\\
\hline

\multicolumn{13}{|c|}{\textbf{(a) BASE}} \\ \hline	
BASE	&	27.8$\times$	&	ST	&	\gradientab{	90.18 	}	90.18 	&	\gradientac{	90.68 	}	90.68	&	\gradientad{	86.82	}	86.82	&	\gradientae{	93.41	}	93.41	&	\gradientaa{	90.27	}	90.27 	$\pm$	1.16 	&	\gradientbb{	82.82	}	82.82	&	\gradientbc{	83.22	}	83.22	&	\gradientbd{	77.64	}	77.64	&	\gradientbe{	87.84	}	87.84	&	\gradientba{	82.88 	}	82.88 	$\pm$	1.67	\\ 

BASE	&	27.8		&	JT	&	\gradientab{	89.42 	}	89.42 	&	\gradientac{	89.89	}	89.89	&	\gradientad{	92.96	}	92.96	&	\gradientae{	94.24	}	94.24	&	\gradientaa{	91.63	}	91.63 	$\pm$	0.42 	&	\gradientbb{	81.68	}	81.68	&	\gradientbc{	82.07	}	82.07	&	\gradientbd{	87.03	}	87.03	&	\gradientbe{	89.36	}	89.36	&	\gradientba{	85.04 	}	85.04 	$\pm$	0.64	\\ 

 \hline
\multicolumn{13}{|c|}{\textbf{(b) Our Method}} \\
\hline
\modelname{}	&	28.5		&	MAT	&	\gradientab{	90.29 	}	90.29 	&	\gradientac{	90.78	}	90.78	&	\gradientad{	93.22	}	\textbf{93.22}	&	\gradientae{	95.53	}	\textbf{95.53}	&	\gradientaa{	92.45	}	\textbf{92.45 	$\pm$	0.65} 	&	\gradientbb{	82.99	}	82.99	&	\gradientbc{	83.41	}	83.41	&	\gradientbd{	87.80	}	\textbf{87.80}	&	\gradientbe{	91.57	}	\textbf{91.57}	&	\gradientba{	86.44 	}	\textbf{86.44 	$\pm$	0.94}	\\

\hline
\multicolumn{13}{|c|}{\textbf{(c) Other Data-efficient MIS ViTs}} \\ 	
\hline

SwinUnet	&	41.4$\times$	&	ST	&	\gradientab{	89.25 	}	89.25 	&	\gradientac{	90.69	}	90.69	&	\gradientad{	88.58	}	88.58	&	\gradientae{	94.13	}	94.13	&	\gradientaa{	90.6625	}	90.66 	$\pm$	0.87 	&	\gradientbb{	81.51	}	81.51	&	\gradientbc{	83.25	}	83.25	&	\gradientbd{	80.40	}	80.40	&	\gradientbe{	89.00	}	89.00	&	\gradientba{	83.54 	}	83.54 	$\pm$	1.27	\\ 
 
SwinUnet	&	41.4		&	JT	&	\gradientab{	89.64 	}	89.64 	&	\gradientac{	90.40	}	90.40	&	\gradientad{	92.98	}	92.98	&	\gradientae{	94.86	}	94.86	&	\gradientaa{	91.97	}	91.97 	$\pm$	0.30 	&	\gradientbb{	81.98	}	81.98	&	\gradientbc{	82.80	}	82.80	&	\gradientbd{	87.08	}	87.08	&	\gradientbe{	90.33	}	90.33	&	\gradientba{	85.55 	}	85.55 	$\pm$	0.50	\\
\hline

UTNet	&	10.0$\times$	&	ST	&	\gradientab{	89.74 	}	89.74 	&	\gradientac{	90.01	}	90.01	&	\gradientad{	88.13	}	88.13	&	\gradientae{	93.23	}	93.23	&	\gradientaa{	90.28	}	90.28 	$\pm$	0.62 	&	\gradientbb{	82.16	}	82.16	&	\gradientbc{	82.13	}	82.13	&	\gradientbd{	79.87	}	79.87	&	\gradientbe{	87.60	}	87.60	&	\gradientba{	82.94 	}	82.94 	$\pm$	0.82	\\ 	
 
UTNet	&	10.0		&	JT	&	\gradientab{	90.24 	}	90.24 	&	\gradientac{	89.85	}	89.85	&	\gradientad{	92.06	}	92.06	&	\gradientae{	94.75	}	94.75	&	\gradientaa{	91.72	}	91.72 	$\pm$	0.63 	&	\gradientbb{	82.92	}	82.92	&	\gradientbc{	82.00	}	82.00	&	\gradientbd{	85.66	}	85.66	&	\gradientbe{	90.17	}	90.17	&	\gradientba{	85.19 	}	85.19 	$\pm$	0.96	\\ \hline	

BAT	&	32.2$\times$	&	ST	&	\gradientab{	90.45 	}	\textbf{90.45} 	&	\gradientac{	90.56	}	90.56	&	\gradientad{	90.78	}	90.78	&	\gradientae{	94.72	}	94.72	&	\gradientaa{	91.63	}	91.63 	$\pm$	0.68 	&	\gradientbb{	83.04	}	83.04	&	\gradientbc{	82.97	}	82.97	&	\gradientbd{	83.66	}	83.66	&	\gradientbe{	90.03	}	90.03	&	\gradientba{	84.92 	}	84.92 	$\pm$	1.01	\\ 

BAT	&	32.2		&	JT	&	\gradientab{	90.06 	}	90.06 	&	\gradientac{	90.06	}	90.06	&	\gradientad{	92.66	}	92.66	&	\gradientae{	93.53	}	93.53	&	\gradientaa{	91.58	}	91.58 	$\pm$	0.33 	&	\gradientbb{	82.44	}	82.44	&	\gradientbc{	82.18	}	82.18	&	\gradientbd{	86.48	}	86.48	&	\gradientbe{	88.11	}	88.11	&	\gradientba{	84.80 	}	84.80 	$\pm$	0.53	\\ \hline	

TransFuse	&	26.3$\times$	&	ST	&	\gradientab{	90.43 	}	90.43 	&	\gradientac{	91.04	}	\textbf{91.04}	&	\gradientad{	91.37	}	91.37	&	\gradientae{	94.93	}	94.93	&	\gradientaa{	91.94	}	91.94 	$\pm$	0.67 	&	\gradientbb{	83.18	}	\textbf{83.18}	&	\gradientbc{	83.86	}	\textbf{83.86}	&	\gradientbd{	84.91	}	84.91	&	\gradientbe{	90.44	}	90.44	&	\gradientba{	85.60 	}	85.60 	$\pm$	0.95	\\ 
 
TransFuse	&	26.3		&	JT	&	\gradientab{	90.03 	}	90.03 	&	\gradientac{	90.48	}	90.48	&	\gradientad{	92.54	}	92.54	&	\gradientae{	95.14	}	95.14	&	\gradientaa{	92.05	}	92.05 	$\pm$	0.36 	&	\gradientbb{	82.56	}	82.56	&	\gradientbc{	82.97	}	82.97	&	\gradientbd{	86.50	}	86.50	&	\gradientbe{	90.85	}	90.85	&	\gradientba{	85.72 	}	85.72 	$\pm$	0.56	\\ \hline	
 
Swin UNETR	&	25.1$\times$	&	ST	&	\gradientab{	90.29 	}	90.29 	&	\gradientac{	90.95	}	90.95	&	\gradientad{	91.10	}	91.10	&	\gradientae{	94.45	}	94.45	&	\gradientaa{	91.70	}	91.70 	$\pm$	0.51 	&	\gradientbb{	82.93	}	82.93	&	\gradientbc{	83.69	}	83.69	&	\gradientbd{	84.16	}	84.16	&	\gradientbe{	89.59	}	89.59	&	\gradientba{	85.09 	}	85.09 	$\pm$	0.79	\\ 
 
Swin UNETR	&	25.1		&	JT	&	\gradientab{	89.81 	}	89.81 	&	\gradientac{	90.87	}	90.87	&	\gradientad{	92.29	}	92.29	&	\gradientae{	94.73	}	94.73	&	\gradientaa{	91.93	}	91.93 	$\pm$	0.29 	&	\gradientbb{	82.21	}	82.21	&	\gradientbc{	83.58	}	83.58	&	\gradientbd{	86.10	}	86.10	&	\gradientbe{	90.11	}	90.11	&	\gradientba{	85.50 	}	85.50 	$\pm$	0.44	\\ 

\hline
\multicolumn{13}{|c|}{\textbf{(d) Other Multi-domain Learning Methods}} \\
\hline

Rundo et al.	&	28.2	&	MAT	&	\gradientab{	89.43 	}	89.43 	&	\gradientac{	89.46	}	89.46	&	\gradientad{	92.62	}	92.62	&	\gradientae{	94.68	}	94.68	&	\gradientaa{	91.55	}	91.55 	$\pm$	0.64 	&	\gradientbb{	81.73	}	81.73	&	\gradientbc{	81.40	}	81.40	&	\gradientbd{	86.71	}	86.71	&	\gradientbe{	90.12	}	90.12	&	\gradientba{	84.99 	}	84.99 	$\pm$	0.90	\\ \hline	

Wang et al.	&	28.1		&	MAT	&	\gradientab{	89.46 	}	89.46 	&	\gradientac{	89.62	}	89.62	&	\gradientad{	92.62	}	92.62	&	\gradientae{	94.47	}	94.47	&	\gradientaa{	91.55	}	91.55 	$\pm$	0.54 	&	\gradientbb{	81.79	}	81.79	&	\gradientbc{	81.59	}	81.59	&	\gradientbd{	86.71	}	86.71	&	\gradientbe{	89.76	}	89.76	&	\gradientba{	84.96 	}	84.96 	$\pm$	0.74	\\ \hline	
 
BASE$^{\dagger}$	&	27.8(.02$\times$)		&	MAT	&	\gradientab{	90.22 	}	90.22 	&	\gradientac{	90.61	}	90.61	&	\gradientad{	93.69	}	\textbf{93.69}	&	\gradientae{	95.55	}	95.55	&	\gradientaa{	92.52	}	92.52 	$\pm$	0.45 	&	\gradientbb{	82.91	}	82.91	&	\gradientbc{	83.14	}	83.14	&	\gradientbd{	88.28	}	\textbf{88.28}	&	\gradientbe{	91.58	}	91.58	&	\gradientba{	86.48 	}	86.48 	$\pm$	0.74	\\ \hline	
 
\modelname{}$^{\dagger}$	&	28.6(.02$\times$)		&	MAT	&	\gradientab{	90.24 	}	\textbf{90.24} 	&	\gradientac{	90.71	}	\textbf{90.71}	&	\gradientad{	93.38	}	93.38	&	\gradientae{	95.9	}	\textbf{95.90}	&	\gradientaa{	92.56	}	\textbf{92.56 $\pm$ 0.52} 	&	\gradientbb{	82.97	}	\textbf{82.97}	&	\gradientbc{	83.31	}	\textbf{83.31}	&	\gradientbd{	88.06	}	88.06	&	\gradientbe{	92.19	}	\textbf{92.19}	&	\gradientba{	86.64 	}	\textbf{86.64 $\pm$ 0.76} 		\\ \hline
\end{tabular}}
\end{table*}

\section{Experiments}
\noindent\textbf{Datasets and Evaluation Metrics:} We study 4 skin lesion segmentation data-sets collected from varied sources: ISIC 2018 (ISIC)~\cite{codella2019skin}, Dermofit Image Library (DMF)~\cite{ballerini2013color}, Skin Cancer Detection (SCD)~\cite{glaister2013msim}, and PH2~\cite{mendoncca2013ph}, which contain 2594, 1300, 206, and 200 samples, respectively. To facilitate a fairer performance comparison across datasets, as in~\cite{bayasi2021culprit}, we only use the 1212 images from DMF that exhibited similar lesion conditions as those in other datasets. We perform 5-fold cross-validation and utilize Dice and IOU metrics for evaluation as~\cite{wang2021boundary}. 

\noindent\textbf{Implementation Details:} We conduct 3 training paradigms: separate (ST), joint (JT), and multi-domain adaptive training (MAT), described in Section~\ref{section:introduction}, to train all the models from scratch on the skin datasets. Images are resized to $256 \times 256$ and then augmented through random scaling, shifting, rotation, flipping, Gaussian noise, and brightness and contrast changes. The encoding transformer blocks' channel dimensions are [64, 128, 320, 512] (Fig.~\ref{fig:model_overview}-a). We use two transformer layers in each transformer block and set the number of heads in MHSA to 8. The hidden dimensions of the CNN bridge and auxiliary peers are 1024 and 512. We deploy models on a single TITAN V GPU and train them for 200 epochs with the AdamW~\cite{loshchilov2017decoupled} optimizer, a batch size of 16, ensuring 4 samples from each dataset, and an initial learning rate of $1 \times 10^{-4}$, which changes through a linear decay scheduler whose step size is 50 and decay factor $\gamma=0.5$.

\noindent\textbf{Comparing Against BASE:} In Table~\ref{table:results}-a,b, compared with BASE in ST, BASE in JT improves the segmentation performance on small datasets (PH2 and SCD) but at the expense of diminished performance on larger datasets (ISIC and DMF). This is expected given the non-negligible inter-domain heterogeneity between skin lesion datasets, as found by Bayasi et al.~\cite{bayasi2022boosternet}. The above results demonstrate that shared knowledge in related domains facilitates training a ViT on small datasets while, without a well-designed multi-domain algorithm, causing negative knowledge transfer (NKT) due to inter-domain heterogeneity, i.e., the model's performance decreases on other datasets. Meanwhile, MDViT fits all the domains without NKT and outperforms BASE in ST by a large margin; significantly increasing Dice and IOU on SCD by 6.4\% and 10.16\%, showing that \modelname{} smartly selects valuable knowledge when given data from a certain domain. Additionally, \modelname{} outperforms BASE in JT across all the domains, with average improvements of 0.82\%  on Dice and 1.4\% on IOU.

\begin{figure}[t]
\centering
\includegraphics[width=1\linewidth]{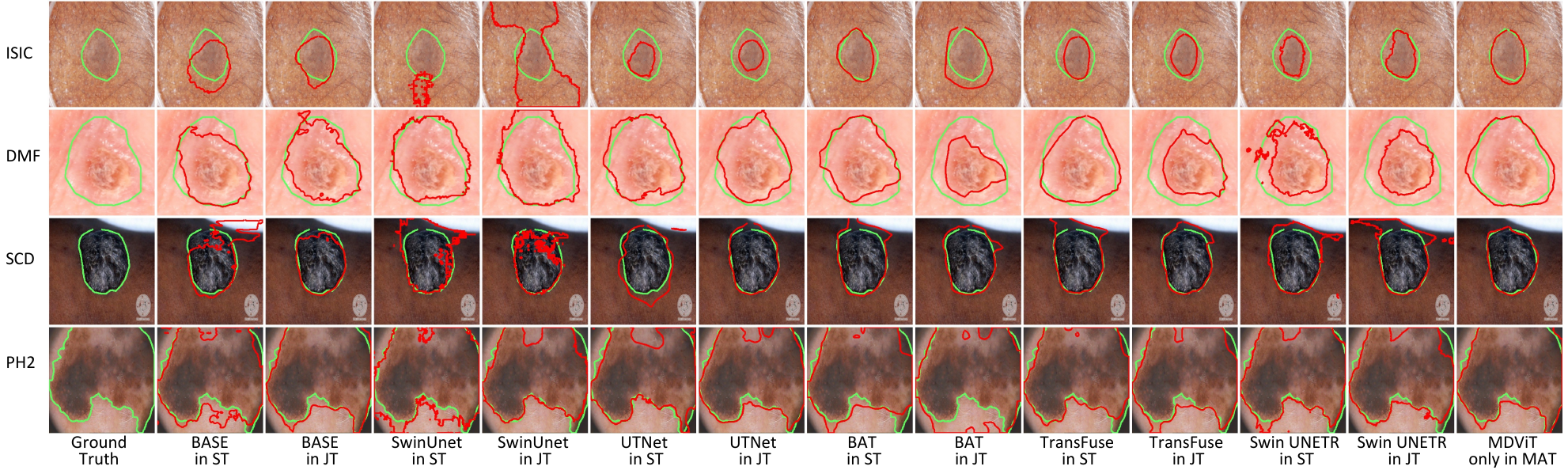}
\caption{Visual result comparison of MDViT, BASE and SOTA data-efficient MIS ViTs in ST and JT training paradigms on four datasets. The green and red contours present the ground truth and segmentation results, respectively.}
\label{fig:visualization}
\end{figure}

\noindent\textbf{Comparing Against State-of-the-Art (SOTA) Methods:} We conduct experiments on SOTA data-efficient MIS ViTs and multi-domain learning methods. Previous MIS ViTs mitigated the data-hunger in one dataset by adding inductive bias, e.g., SwinUnet~\cite{cao2023swin}, UTNet~\cite{gao2021utnet}, BAT~\cite{wang2021boundary}, TransFuse~\cite{zhang2021transfuse}, and Swin UNETR~\cite{tang2022self}. We implement ResNet-34 as the backbone of BAT for fair comparison (similar model size). As illustrated in Table~\ref{table:results}-a,b,c, these SOTA models are superior to BASE in SJ. This is expected since they are designed to reduce data requirements. Nevertheless, in JT, these models also suffer from NKT: They perform better than models in ST on some datasets, like SCD, and worse on others, like ISIC. Finally, \modelname{} achieves the best segmentation performance in average Dice and IOU without NKT and has the best results on SCD and PH2. Fig.~\ref{fig:visualization} shows \modelname{}'s excellent performance on ISIC and DMF and that it achieves the closest results to ground truth on SCD and PH2. More segmentation results are presented in the supplementary material. Though BAT and TransFuse in ST have better results on some datasets like ISIC, they require extra compute resources to train $M$ models as well as an $M$-fold increase in memory requirements. The above results indicate that domain-shared knowledge is especially beneficial for training relatively small datasets such as SCD.

We employ the two fixed-size (i.e., independent of $M$) multi-domain algorithms proposed by Rundo et al.~\cite{rundo2019use} and Wang et al.~\cite{wang2019towards} on BASE. We set the number of parallel SE adapters in~\cite{wang2019towards} to 4. In Table~\ref{table:results}-b,d, \modelname{} outperforms both of them on all the domains, showing the efficacy of \modelname{} and that multi-domain methods built on ViTs might not perform as well as on CNNs. We also apply the domain-specific normalization~\cite{liu2020ms} to BASE and \modelname{} to get BASE$^{\dagger}$ and \modelname{}$^{\dagger}$, respectively. In Table~\ref{table:results}-d, BASE$^{\dagger}$ confronts NKT, which lowers the performance on DMF compared with BASE in ST, whereas \modelname{}$^{\dagger}$ not only addresses NKT but also outperforms BASE$^{\dagger}$ on average Dice and IOU.

\begin{table*}[t]
\centering
\caption{Ablation studies of \modelname{} and experiments of DA's plug-in capability. KD means general knowledge distillation, i.e., we only transfer knowledge from auxiliary peers to the universal network. $^D$ or $^B$ refers to using DeepLabv3's decoder or BASE's decoding layers as auxiliary peers.}
\label{table:ablation}
\resizebox{\textwidth}{!}{
\begin{tabular}{|p{29mm}|p{15mm}|P{9mm}|P{10mm}|P{10mm}|P{10mm}|P{10mm}|P{21mm}|P{10mm}|P{10mm}|P{10mm}|P{10mm}|P{21mm}|}
\hline
\textbf{Model} & \multicolumn{1}{c|}{\textbf{\#Param.}} & \textbf{T} & \multicolumn{5}{c|}{Dice $\uparrow$}&  \multicolumn{5}{c|}{IOU $\uparrow$} \\
\cline{4-13}
~ & \multicolumn{1}{c|}{\textbf{(M)}} &  ~ & ISIC & DMF & SCD & PH2 & avg $\pm$ std  & ISIC & DMF & SCD & PH2 & avg $\pm$ std \\
\hline

\multicolumn{13}{|c|}{\textbf{(a) Plug-in Capability of DA}} \\
\hline

DosViT	&	14.6	&	JT	&	\gradientcb{	88.66	}	88.66	&	\gradientcc{	89.72	}	89.72	&	\gradientcd{	90.65	}	90.65	&	\gradientce{	94.26	}	94.26	&	\gradientca{	90.82 	}	90.82 	$\pm$	0.43 	&	\gradientdb{	80.45 	}	80.45 	&	\gradientdc{	81.73 	}	81.73 	&	\gradientdd{	83.29 	}	83.29 	&	\gradientde{	89.26 	}	89.26 	&	\gradientda{	83.68 	}	83.68 	$\pm$	0.68 	\\ \hline

DosViT+DA	&	14.9	&	MAT	&	\gradientcb{	89.22	}	89.22	&	\gradientcc{	89.91	}	89.91	&	\gradientcd{	90.73	}	90.73	&	\gradientce{	94.42	}	94.42	&	\gradientca{	91.07 	}	91.07 	$\pm$	0.32  	&	\gradientdb{	81.28 	}	81.28 	&	\gradientdc{	82.00 	}	82.00 	&	\gradientdd{	83.44 	}	83.44 	&	\gradientde{	89.57 	}	89.57 	&	\gradientda{	84.07 	}	84.07 	$\pm$	0.50 	\\ \hline
 
TransFuse &	26.3	&	JT	&	\gradientcb{	90.03	}	90.03	&	\gradientcc{	90.48	}	90.48	&	\gradientcd{	92.54	}	92.54	&	\gradientce{	95.14	}	95.14	&	\gradientca{	92.05 	}	92.05 	$\pm$	0.36  	&	\gradientdb{	82.56 	}	82.56 	&	\gradientdc{	82.97 	}	82.97 	&	\gradientdd{	86.50 	}	86.50 	&	\gradientde{	90.85 	}	90.85 	&	\gradientda{	85.72 	}	85.72 	$\pm$	0.56	 \\ \hline
 
TransFuse+DA	&	26.9	&	MAT	&	\gradientcb{	90.13	}	90.13	&	\gradientcc{	90.47	}	90.47	&	\gradientcd{	93.62	}	\textbf{93.62}	&	\gradientce{	95.21	}	95.21	&	\gradientca{	92.36 	}	92.36 	$\pm$	0.38 	&	\gradientdb{	82.80 	}	82.80 	&	\gradientdc{	82.94 	}	82.94 	&	\gradientdd{	88.16 	}	\textbf{88.16} 	&	\gradientde{	90.97 	}	90.97 	&	\gradientda{	86.22 	}	86.22 	$\pm$	0.64  	\\ \hline

\multicolumn{13}{|c|}{\textbf{(b) Ablation Study for DA and MKD}} \\
\hline

BASE &	27.8	&	JT	&	\gradientcb{	89.42	}	89.42	&	\gradientcc{	89.89	}	89.89	&	\gradientcd{	92.96	}	92.96	&	\gradientce{	94.24	}	94.24	&	\gradientca{	91.63 	}	91.63 	$\pm$	0.42  	&	\gradientdb{	81.68 	}	81.68 	&	\gradientdc{	82.07 	}	82.07 	&	\gradientdd{	87.03 	}	87.03 	&	\gradientde{	89.36 	}	89.36 	&	\gradientda{	85.04 	}	85.04 	$\pm$	0.64 	\\ \hline
	
BASE+DA	&	28.5	&	MAT	&	\gradientcb{	89.96	}	89.96	&	\gradientcc{	90.66	}	90.66	&	\gradientcd{	93.36	}	93.36	&	\gradientce{	95.46	}	95.46	&	\gradientca{	92.36 	}	92.36 	$\pm$	0.51  	&	\gradientdb{	82.52 	}	82.52 	&	\gradientdc{	83.24 	}	83.24 	&	\gradientdd{	87.98 	}	87.98 	&	\gradientde{	91.43 	}	91.43 	&	\gradientda{	86.29 	}	86.29 	$\pm$	0.72  	\\ \hline
 
BASE+MKD	&	27.8	&	JT	&	\gradientcb{	89.27	}	89.27	&	\gradientcc{	89.53	}	89.53	&	\gradientcd{	92.66	}	92.66	&	\gradientce{	94.83	}	94.83	&	\gradientca{	91.57 	}	91.57 	$\pm$	0.53  	&	\gradientdb{	81.45 	}	81.45 	&	\gradientdc{	81.49 	}	81.49 	&	\gradientdd{	86.81 	}	86.81 	&	\gradientde{	90.42 	}	90.42 	&	\gradientda{	85.04 	}	85.04 	$\pm$	0.74	 \\ \hline
 
BASE+DA+KD	&	28.5	&	MAT	&	\gradientcb{	90.03	}	90.03	&	\gradientcc{	90.59	}	90.59	&	\gradientcd{	93.26	}	93.26	&	\gradientce{	95.63	}	 \textbf{95.63}	&	\gradientca{	92.38 	}	92.38 	$\pm$	0.39 	&	\gradientdb{	82.67 	}	82.67 	&	\gradientdc{	83.12 	}	83.12 	&	\gradientdd{	87.85 	}	87.85 	&	\gradientde{	91.72 	}	\textbf{91.72} 	&	\gradientda{	86.34 	}	86.34 	$\pm$	0.51	 \\ \hline
 
\multicolumn{13}{|c|}{\textbf{(c) Ablation Study for Auxiliary Peers}} \\
\hline

\modelname{}$^D$	&	28.5	&	MAT	&	\gradientcb{	89.64	}	89.64	&	\gradientcc{	90.25	}	90.25	&	\gradientcd{	92.24	}	92.24	&	\gradientce{	95.36	}	95.36	&	\gradientca{	91.87 	}	91.87 	$\pm$	0.45 	&	\gradientdb{	82.10 	}	82.10 	&	\gradientdc{	82.55 	}	82.55 	&	\gradientdd{	86.12 	}	86.12 	&	\gradientde{	91.24 	}	91.24 	&	\gradientda{	85.50 	}	85.50 	$\pm$	0.67 	\\ \hline

\modelname{}$^B$	&	28.5	&	MAT	&	\gradientcb{	90.03	}	90.03	&	\gradientcc{	90.73	}	90.73	&	\gradientcd{	92.72	}	92.72	&	\gradientce{	95.32	}	95.32	&	\gradientca{	92.20 	}	92.20 	$\pm$	0.50 	&	\gradientdb{	82.66 	}	82.66 	&	\gradientdc{	83.35 	}	83.35 	&	\gradientdd{	87.01 	}	87.01 	&	\gradientde{	91.17 	}	91.17 	&	\gradientda{	86.05 	}	86.05 	$\pm$	0.70 	\\ \hline
 
\modelname{} &	28.5	&	MAT	&	\gradientcb{	90.29	}	\textbf{90.29}	&	\gradientcc{	90.78	}	\textbf{90.78}	&	\gradientcd{	93.22	}	93.22	&	\gradientce{	95.53	}	95.53	&	\gradientca{	92.45 	}	\textbf{92.45 	$\pm$	0.65} & 	\gradientdb{	82.99 	}	\textbf{82.99} 	&	\gradientdc{	83.41 	}	\textbf{83.41} 	&	\gradientdd{	87.80 	}	87.80 	&	\gradientde{	91.57 	}	91.57 	&	\gradientda{	86.44 	}	\textbf{86.44 	$\pm$	0.94} 	 \\ \hline
 
\end{tabular}}
\end{table*}

\noindent\textbf{Ablation Studies and Plug-in Capability of DA:} We conduct ablation studies to demonstrate the efficacy of DA, MKD, and auxiliary peers. Table~\ref{table:ablation}-b reveals that using one-direction knowledge distillation (KD) or either of the critical components in \modelname{}, i.e., DA or MKD, but not together, could not achieve the best results. Table~\ref{table:ablation}-c exemplifies that, for building the auxiliary peers, our proposed MLP architecture is more effective and has fewer parameters ($1.6$M) than DeepLabv3's decoder~\cite{chen2017rethinking} ($4.7$M) or BASE's decoding layers (10.8M). Finally, we incorporate DA into two ViTs: TransFuse and DosViT (the latter includes the earliest ViT encoder~\cite{dosovitskiy2020image} and a DeepLabv3's decoder). As shown in Table~\ref{table:ablation}-a,b, DA can be used in various ViTs but is more advantageous in \modelname{} with more transformer blocks in the encoding and decoding process.

\section{Conclusion}
We propose a new algorithm to alleviate vision transformers (ViTs)' data-hunger in small datasets by aggregating valuable knowledge from multiple related domains. We constructed \modelname{}, a robust multi-domain ViT leveraging novel domain adapters (DAs) for negative knowledge transfer mitigation and mutual knowledge distillation (MKD) for better representation learning. \modelname{} is non-scalable, i.e., has a fixed model size at inference time even as more domains are added. The experiments on 4 skin lesion segmentation datasets show that \modelname{} outperformed SOTA data-efficient medical image segmentation ViTs and multi-domain learning methods. Our ablation studies and application of DA on other ViTs show the effectiveness of DA and MKD and DA's plug-in capability.

\bibliographystyle{splncs04}
\bibliography{mybibliography}

\end{document}